\documentclass{article}

\usepackage{microtype}
\usepackage{graphicx}
\usepackage{booktabs} % for professional tables
\usepackage{enumitem}
\usepackage{subcaption}
\usepackage{hyperref}

\usepackage[accepted]{icml2025}

\usepackage{amsmath}
\usepackage{amssymb}
\usepackage{mathtools}
\usepackage{amsthm}

\usepackage[capitalize,noabbrev]{cleveref}

\theoremstyle{plain}

\theoremstyle{definition}

\theoremstyle{remark}

\usepackage[textsize=tiny]{todonotes}

\icmltitlerunning{}

\begin{document}

\twocolumn[
\icmltitle{Revisiting the Relationship between Adversarial and Clean Training: Why Clean Training Can Make Adversarial Training Better}




\begin{icmlauthorlist}
\icmlauthor{MingWei Zhou and Xiaobing Pei}{}
%\icmlauthor{zmw}{sch}
\end{icmlauthorlist}

\icmlkeywords{Machine Learning, ICML}

\vskip 0.3in
]




\begin{abstract}
Adversarial training (AT) is an effective technique for enhancing  adversarial robustness, but it usually comes at the cost of a decline in generalization ability. Recent studies have attempted to use clean  training to assist adversarial training, yet there are contradictions  among the conclusions. We comprehensively summarize the representative  strategies and, with a focus on the multi - view hypothesis, provide a  unified explanation for the contradictory phenomena among different  studies. In addition, we conduct an in - depth analysis of the knowledge combinations transferred from clean - trained models to adversarially - trained models in previous studies, and find that they can be divided  into two categories: reducing the learning difficulty and providing  correct guidance. Based on this finding, we propose a new idea of  leveraging clean training to further improve the performance of advanced AT methods.We reveal that the problem of generalization degradation faced by AT  partly stems from the difficulty of adversarial training in learning  certain sample features, and this problem can be alleviated by making  full use of clean training.
\end{abstract}

\section{Introduction}
Deep-learning models are vulnerable to adversarial examples.Adversarial  examples are visually indistinguishable from normal samples but can  mislead models into making wrong judgments \cite{a7,a8}. To address this  challenge, various adversarial defense algorithms have been proposed  \cite{a9,a10,a11,a12,a13,a14,a15,a21}, among which adversarial training \cite{a16b20}  has been proven to be the most effective method for enhancing the  robustness of neural networks \cite{a22,a23}. In a large number of studies on  adversarial training, models with high robustness often come at the cost of sacrificing natural accuracy \cite{a16b20,b47,b16}.

To mitigate this challenge, numerous studies have explored from different  perspectives. For example, some studies added additional data \cite{k12,k13},  some modified the model training paradigm \cite{b86}, and some relaxed the  perturbation magnitude \cite{k10,k11}. In addition, the label design for  adversarial training has become a highly concerned and promising  research perspective. Compared with other perspectives, it has a lower  computational cost. Moreover, it has great  potential in revealing the essence of why adversarial examples are  difficult to learn.

To strengthen adversarial training (AT), the label-design methods for  adversarial training usually aim to generate more accurate sample labels from different perspectives. LBGAT believes that the adversarial training model should inherit the classification boundary  of the clean - trained model and changes the label to the prediction of  the co - trained clean model \cite{b50a25}. LTD  believes that  the commonly used labels (i.e., one - hot vectors) hinder the learning  process of image recognition and modifies the label to the prediction of the clean model after adding a temperature coefficient \cite{b68a26}. KD-AT-Auto  believes that the original labels are noisy for  adversarial training and changes the label to the interpolation between  the prediction of the adversarial model after adding a temperature  coefficient and the original label \cite{label}. HAT believes that the original labels may cause the adversarial training to lead to  an unwarranted increase in the margin along certain adversarial  directions and sets the label of the adversarial sample with double  perturbation as the category that is most easily perturbed \cite{b85}. There  are also some methods that distill the capabilities of larger  adversarial training models into smaller ones. These methods, such as  ARD\cite{b69}, IAD\cite{b70}, and RSLAD\cite{b71}, mainly believe that the predictions  of the adversarial models should be aligned with those of larger  adversarial models, and all consider that the predictions of clean  training as labels are harmful to adversarial training.

Although these studies provide various strategies for the label design of  adversarial training, their underlying ideas are diverse and sometimes  even contradictory. For example, the clean - sample predictions focused  on in LBGAT and LTD are exactly what ARD\cite{b69}, IAD\cite{b70}, and RSLAD\cite{b71}  try to avoid. To eliminate the confusion caused by these differences in  future research, it is necessary to systematically summarize and  uniformly explain the label design of adversarial training to promote  the development of this field.

In this paper, we reveal that the problem of the degradation of  generalization ability faced by AT partly stems from the difficulty of  adversarial training in learning certain sample features. And we propose a unified framework to summarize the label - design strategies used in  representative studies by dividing the label design into two key parts:  providing correct guidance and reducing the learning difficulty of the  model. Our research shows that providing correct guidance and reducing  the learning difficulty interact with each other and need to be designed in coordination. And the ideas of existing label - design strategy  studies can be interpreted as different ways of dealing with providing  correct guidance and reducing the learning difficulty. We also prove the potential for further improvement in the topic of label - design  strategies through the roles of providing correct guidance and reducing  the learning difficulty.

By further exploring the two major steps of providing correct guidance and reducing the learning difficulty, we propose a novel label-design  method. This method emphasizes that clean training can provide correct  guidance for adversarial training, but various ways are needed to reduce the learning difficulty during the guidance process. By applying this  processing method to adversarial training, we better alleviate the  problem of the decline in the accuracy of adversarial training from the  perspective of label design. Specifically, our contributions are  summarized as follows:

\begin{itemize}[nosep]
	\item  We propose a new explanation for the decline in the natural  generalization ability of adversarial training models, revealing that  the problem of the degradation of generalization ability faced by AT  partly stems from the difficulty of adversarial training in learning  certain sample features.
	\item We conduct a systematic analysis of the label - design methods in  adversarial training, divide the label design into two key parts:  providing correct guidance and reducing the learning difficulty, to  summarize the label - design strategies used in representative studies.  And we thus propose a novel idea for the label design of adversarial  training, fully exploring the potential for improving adversarial  training through label design.
	\item We conduct extensive experiments on different classification models  and datasets to prove the effectiveness of our method. The results show  that compared with related techniques, our improved method significantly improves the normal accuracy and robust accuracy on benchmark datasets.
\end{itemize}
\section{Related Work}

Despite various defense methods, a series of studies \cite{b60,b61} have revealed that adversarial training remains one of the best defense strategies. The basic idea behind adversarial training is to solve a min-max optimization problem, as shown in Equation (1):
\abovedisplayskip=0pt\relax
\belowdisplayskip=0pt\relax
\begin{equation}\label{eq:single}
	\min_{\theta} \rho(\theta),~where~\rho(\theta) = \mathbb{E}_{(x,y) \sim \mathcal{D}} \left[ \max_{\delta \in S} L(\theta, x + \delta, y) \right]\end{equation}
Where ${\theta}$ represents the parameters of the adversarial model, ${\delta}$ represents the perturbation size, ${x}$ represents clean samples, ${y}$ represents the corresponding labels of ${x}$, ${L}$ represents the loss function, and ${\rho(\theta)}$ represents the distribution difference between adversarial samples and clean samples.

Methods such as PGDAT \cite{a16b20}, TRADES \cite{b47}, and MART \cite{b16}  have successfully enhanced the robustness of deep models. However,  compared with cleanly - trained models, the natural accuracy of  adversarial models still significantly declines. Recently, numerous  methods, including LBGAT \cite{b50a25}, LTD \cite{b68a26}, KD-AT-Auto \cite{label}, HAT \cite{b85},  and a series of adversarial training distillation methods such as ARD\cite{b69}, IAD\cite{b70}, and RSLAD\cite{b71}, have demonstrated the effectiveness of label  design in adversarial training (AT). Nevertheless, some of the research  conclusions among them are contradictory.

In this paper, we reveal that the problem of the decline in generalization ability faced by AT partly stems from the difficulty of adversarial  training in learning certain sample features. Previous studies have  played a role in learning these sample features from two aspects:  reducing the learning difficulty and providing correct guidance.  Therefore, we further explore the truly correct guidance and methods to  reduce the learning difficulty, and propose CKTAT.

\section{Hard-to-learn single-view samples}
\subsection{Preliminaries}
\subsubsection{Multi-View Hypothesis}
The multi-view data hypothesis be proposed in work \cite{b18}.Consider a binary classification problem with four features ${v_1, v_2, v_3, v_4}$. The first two features correspond to the first class label, and the last two features correspond to the second class label. In the data distribution:

When belonging to the first class label, we have:
\abovedisplayskip=0pt\relax
\belowdisplayskip=0pt\relax
\begin{equation}\label{eq:single}{\begin{cases}  
			
			\text{Both } v_1 \text{ and } v_2 \text{ appear with weight } 1, & \text{accounting for 60\%} \\  
			
			\text{Only } v_1 \text{ appears with weight } 1, & \text{accounting for 20\%} \\  
			
			\text{Only } v_2 \text{ appears with weight } 1, & \text{accounting for 20\%}  
			
\end{cases}}\end{equation}

When belonging to the second class label, we have:
\begin{equation}\label{eq:single}{\begin{cases}  
			
			\text{Both } v_3 \text{ and } v_4 \text{ appear with weight } 1, & \text{accounting for 60\%} \\  
			
			\text{Only } v_3 \text{ appears with weight } 1, & \text{accounting for 20\%} \\  
			
			\text{Only } v_4 \text{ appears with weight } 1, & \text{accounting for 20\%}  
			
\end{cases}}\end{equation}
We call 60\% of the data multi-view data: these data have multiple features that can be used for correct classification. The remaining 40\% of the data is called single-view data: lacking some features of the correct label.
\subsubsection{The Dense Mixtures of Network's Weight}
Recently, Zhu Zeyuan et al.\cite{b17} pointed that the reason for the existence of adversarial examples is that the neural network's weight vectors accumulate dense mixtures that are not robust to adversarial perturbations during training. 

Zhu Zeyuan et al. \cite{b17} assumed a sparse coding model for data generation, where each input data ${x}$ takes the form ${x = Mz +\xi}$ , with ${z}$ being a sparse vector and ${\xi}$ being noise. In this context, Zhu Zeyuan et al. \cite{b17} pointed out that the hidden weights of the ${i}$th hidden neuron at iteration ${t}$ can be expressed as shown in Equation (4):
\begin{equation}\label{eq:single}{\omega_{i}^{(t)} = \Theta(1)M_{j} + {\sum_{j^{'} \neq j}\left\lbrack \Theta\left( \frac{k}{d} \right)M_{j^{'}} \right\rbrack}}\end{equation}
Where ${\omega}_i^{t}$ represents the hidden weights of the ${i}$th hidden neuron at iteration ${t}$, ${\theta}$ represents the asymptotically tight bound, ${M_j}$ represents the learned feature weights for feature ${j}$ in the data, ${k}$ represents the number of image classification categories, and ${d}$ represents the dimensionality of ${M}$. ${M_{j^{'}}}$ represents the learned feature weights for other directional features in the data.

After clean training, the neural network can learn a large portion of robust features ${\Theta(1)M_{j}}$, along with some small dense mixtures ${m = {\sum_{j^{'} \neq j}\left\lbrack \Theta\left( \frac{k}{d} \right)M_{j^{'}} \right\rbrack}}$. Thus, Zhu Zeyuan et al. \cite{b17} pointed out that Equation (5) is likely to hold:
\begin{equation}\label{eq:single}{\left\lfloor \left\langle {m,x} \right\rangle \right\rfloor \leq \overset{\sim}{O}\left( \frac{k}{d}\left\| x \right\|_{2} \right)}\end{equation}

Where ${\left\lfloor \left\langle {m,x} \right\rangle \right\rfloor}$represents the absolute value of the inner product between ${m}$ and ${x}$,${\overset{\sim}{O}}$ is the abbreviation of ${O \left( g(n){log}_{k}n \right))}$, ${O}$ represents the asymptotically tight upper bound, ${k}$ is the number of image classification categories, ${d}$ is the dimensionality of ${M}$.

When a perturbation ${\delta}$ is added to the input data \( x \) along the direction of ${\sum_{j^{'} \neq j}\left\lbrack \Theta\left( \frac{k}{d} \right)M_{j^{'}} \right\rbrack}$, that is, ${\delta \propto {\sum_{j^{'} \neq j}\left\lbrack \Theta\left( \frac{k}{d} \right)M_{j^{'}} \right\rbrack}}$, where ${\propto}$ indicates proportionality, Zhu Zeyuan et al. \cite{b17} pointed out that Equation (6) holds:
\begin{equation}\label{eq:single}{\left\lfloor \left\langle {m,\delta} \right\rangle \right\rfloor = \Omega\left( \frac{k}{\sqrt{d}}\left\| \delta \right\|_{2} \right)}\end{equation}
Where ${\left\lfloor \left\langle {m,\delta} \right\rangle \right\rfloor}$ represents the absolute value of the inner product between \( m \) and ${\delta}$,${\Omega}$ represents the asymptotically tight lower bound, \( k \) is the number of image classification categories, \( d \) is the dimensionality of \( M \).

\subsection{Analysis of feature learning for single-view samples}
We find that compared to clean training, adversarial training has a high probability of failing to learn some features in single-view data.The specific analysis is as follows(For the definitions and specific meanings of the following various symbols, please refer to the appendix.):

Lemma B.1 is first given to explain the model's learning process for features:

\textbf{Lemma B.1} \cite{b18} (Model's Feature Learning Process): For each label $i~ \in ~\lbrack k\rbrack$, one of the correlation weights of \( v_{i,1} \) or \( v_{i,2} \) will be learned first by model \( M \).

Based on Lemma B.1, subsequent analysis assumes that model \( M\)  first learn the feature correlation weight of \( v_{y,j} \). Next, we analyze the impact on the clean training error of \( M_{clean} \) when it starts learning the feature correlation weight of \( v_{y,3-j} \), and the impact on the robust training error of\(  M_{adv} \) when it begins learning the feature correlation weight of \( v_{y,3-j} \).

To understand the impact on training error, we need to analyze the relationship between feature correlation weights and outputs. According to \cite{b18}, ideally, the feature correlation weight of \( v_{y,j}  \)learned by model \( M \) can perfectly classify clean data containing the \( v_{y,j}  \) feature.By combining formula (4) and formula (5) in main text, we can obtain the output of the \( v_{y,j}  \) feature-related weights learned by model \( M \) for the \( v_{y,j}  \) feature in clean sample data, as well as the output of the \( v_{y,j}  \) feature-related weights learned by model \( M  \) for the adversarial perturbation $\sigma_{v_{y,{j}}}$ targeting the \( v_{y,j}  \) feature. 
The learned weights related to the feature \( v_{y,j}  \) by Model \( M \) have the following judgments for the feature \( v_{y,j}  \) in a clean sample \( X_i \): $O_{v_{y,{j}}}^{y}\left( {X_{i}}_{v_{y,{j}}} \right) = \Theta(1),O_{v_{y,{j}}}^{l \in ~{\lbrack k\rbrack},l \neq y}\left( {X_{i}}_{v_{y,{j}}} \right) \approx 0$. The learned weights related to the feature \( v_{y,j}  \) by Model \( M \) have the following judgments for the adversarial perturbation \( \sigma_{v_{y,j}} \) targeting the feature \( v_{y,j}  \): $O_{v_{y,{j}}}^{l_{max}}\left( \sigma_{v_{y,{j}}} \right) = k_{1}\left( {\sum_{j^{'} \neq j}\left\lbrack \Theta\left( \frac{k}{d} \right) \right\rbrack} \right)$, where $k_{1} \in \lbrack 0,1\rbrack$, and $l_{max}$ is the category with the maximum value in $O_{v_{y,{j}}}^{l \in ~{\lbrack k\rbrack},l \neq y}\left( \sigma_{v_{y,{j}}} \right)$. According to works \cite{b60,b61}, $O_{v_{y,{j}}}^{l_{max}}\left( \sigma_{v_{y,{j}}} \right) \gg ~O_{v_{y,{j}}}^{l \in ~{\lbrack k\rbrack},l \neq y,l \neq l_{max}}\left( \sigma_{v_{y,{j}}} \right)$. To simplify the problem, this paper only considers the impact of $O_{v_{y,{j}}}^{l_{max}}\left( \sigma_{v_{y,{j}}} \right)$ caused by perturbations. Similarly, combining Equation (4) and Equation (5) in main text, we can obtain the output of the weights related to the feature \( v_{y,3-j} \) subsequently learned by Model \( M \) for the feature \( v_{y,3-j} \) in clean sample data, as well as the output of the weights related to the feature \( v_{y,3-j} \) learned by Model \( M \) for the adversarial perturbation \( \sigma_{v_{y,3-j}} \) targeting the feature \( v_{y,3-j} \). The learned weights related to the feature \( v_{y,3-j} \) by Model M have the following judgments for the feature \( v_{y,3-j} \) in a clean sample \( X_i \): $O_{v_{y,3 - {j}}}^{y}\left( {X_{i}}_{v_{y,3 - {j}}} \right) = \Theta(1),O_{v_{y,3 - {j}}}^{l \in ~{\lbrack k\rbrack},l \neq y}\left( {X_{i}}_{v_{y,3 - {j}}} \right) \approx 0$. At the same time, it is set that the learned weights related to the feature \( v_{y,3-j} \) by Model \( M \) have the following judgments for the adversarial perturbation: $O_{v_{y,3 - {j}}}^{l_{max}}\left( \sigma_{v_{y,3 - {j}}} \right) = k_{2}\left( {\sum_{j^{'} \neq j}\left\lbrack \Theta\left( \frac{k}{d} \right) \right\rbrack} \right)$, where $k_{2} \in \lbrack 0,1\rbrack$, and \( l_{max} \) is the category with the maximum value in $O_{v_{y,3 - {j}}}^{l \in ~{\lbrack k\rbrack},l \neq y}\left( \sigma_{v_{y,3 - {j}}} \right)$. The impact of perturbations is also only considered in terms of $O_{v_{y,3 - {j}}}^{l_{max}}\left( \sigma_{v_{y,3 - {j}}} \right)$. In this paper, we do not consider the reduction of errors caused by noise fitting, that is, we believe that the model can only judge the corresponding features in the data when it learns the relevant features. Before learning the feature \( v_{y,3-j} \), the model cannot make judgments on clean samples \( X_i \) and adversarial sample data \( \hat{X_i} \) that only contain the feature \( v_{y,3-j} \). Therefore, before learning the feature \( v_{y,3-j} \), the clean training error of the model for clean samples \( X_i \) that only contain the feature \( v_{y,3-j}  \) is 1, and the robust training error for adversarial sample data\(  \hat{X_i} \) that only contain the feature \( v_{y,3-j} \) is also 1.

Let \( R_{\text{robust}}^1 \) represent the robust training error of the adversarial model before learning the feature \( v_{y,3-j} \) ,\( R_{\text{clean}}^1 \) represent the clean training error of the clean model before learning the feature \( v_{y,3-j} \), \( R_{\text{robust}}^2  \) represent the robust training error of the adversarial model after learning the feature \( v_{y,3-j} \), and \( R_{\text{clean}}^2 \) represent the clean training error of the clean model after learning the feature \( v_{y,3-j} \).
\begin{equation}
	\label{eq:single}
	\begin{aligned}
		{R_{robust}}^{1} &= \frac{1}{N} \Bigg( \sum_{i = 1}^{0.5\mu N} \Bigg( 2 - \frac{e^{\Theta(1)}}{e^{\Theta(1)} + e^{k_{1}\left(\sum_{j' \neq j} \left[\Theta\left(\frac{k}{d}\right)\right]\right)}} \Bigg) \\
		& \quad + \sum_{i = 1}^{(1 - \mu)N} \Bigg( 1 - \frac{e^{\Theta(1)}}{e^{\Theta(1)} + e^{k_{1}\left(\sum_{j' \neq j} \left[\Theta\left(\frac{k}{d}\right)\right]\right)}} \Bigg) \Bigg)
	\end{aligned}
\end{equation}
To obtain the training error, it is necessary to know the composition of the data, the weight composition of the model, and the final output composition. The relationship between these compositions is shown in Figure 1:

From these, we derive Equations (7), (8), (9), and (10):

\abovedisplayskip=0pt\relax
\belowdisplayskip=0pt\relax
\begin{equation}
	\label{eq:single}
	\begin{aligned}
		{R_{robust}}^{2} &= \frac{1}{N} \Bigg( 
		\sum_{i = 1}^{0.5\mu N} \left( 1 - \frac{e^{\Theta(1)}}{e^{\Theta(1)} + e^{k_{2}\left(\sum_{j' \neq j} \left[\Theta\left(\frac{k}{d}\right)\right]\right)}} \right) \\
		& \quad + \sum_{i = 1}^{0.5\mu N} \left( 1 - \frac{e^{\Theta(1)}}{e^{\Theta(1)} + e^{k_{1}\left(\sum_{j' \neq j} \left[\Theta\left(\frac{k}{d}\right)\right]\right)}} \right) \\
		& \quad + \sum_{i = 1}^{(1 - \mu)N} \left( 1 - \frac{e^{2\Theta(1)}}{e^{2\Theta(1)} + e^{(k_{1} + k_{2})\left(\sum_{j' \neq j} \left[\Theta\left(\frac{k}{d}\right)\right]\right)}} \right) 
		\Bigg)
	\end{aligned}
\end{equation}
\begin{equation}\label{eq:single}{{~R}_{clean}}^{1} = \frac{1}{N}\left( {\sum_{i = 1}^{0.5\mu N}(1)}~ \right)~~\end{equation}
\begin{equation}\label{eq:single}{R_{clean}}^{2} = 0~~~~\end{equation}
The differences in robust training error and clean error before and after learning the \( v_{y,3-j} \) feature are given by Equations (11) and (12):
\begin{equation}
	\label{eq:single}
	\begin{aligned}
		{R_{\text{robust}}}^{2} - {R_{\text{robust}}}^{1} &= \frac{1}{N} \Bigg( 
		\sum_{i = 1}^{0.5\mu N} \left( - \frac{e^{\Theta(1)}}{e^{\Theta(1)} + e^{k_{2}\left(\sum_{j' \neq j} \left[\Theta\left(\frac{k}{d}\right)\right]\right)}} \right) \\
		&\quad + \sum_{i = 1}^{(1 - \mu)N} \left( \frac{e^{\Theta(1)}}{e^{\Theta(1)} + e^{k_{1}\left(\sum_{j' \neq j} \left[\Theta\left(\frac{k}{d}\right)\right]\right)}} \right) \\
		&\quad - \sum_{i = 1}^{(1 - \mu)N} \left( \frac{e^{2\Theta(1)}}{e^{2\Theta(1)} + e^{(k_{1} + k_{2})\left(\sum_{j' \neq j} \left[\Theta\left(\frac{k}{d}\right)\right]\right)}} \right) 
		\Bigg)
	\end{aligned}
\end{equation}
\begin{equation}\label{eq:single}{R_{clean}}^{2} - {R_{clean}}^{1} = - \frac{1}{N}{\sum_{i = 1}^{0.5\mu N}(1)}~~\end{equation}
\begin{figure*}[]  
	
	\centering  
	
	\includegraphics[width=0.7\textwidth]{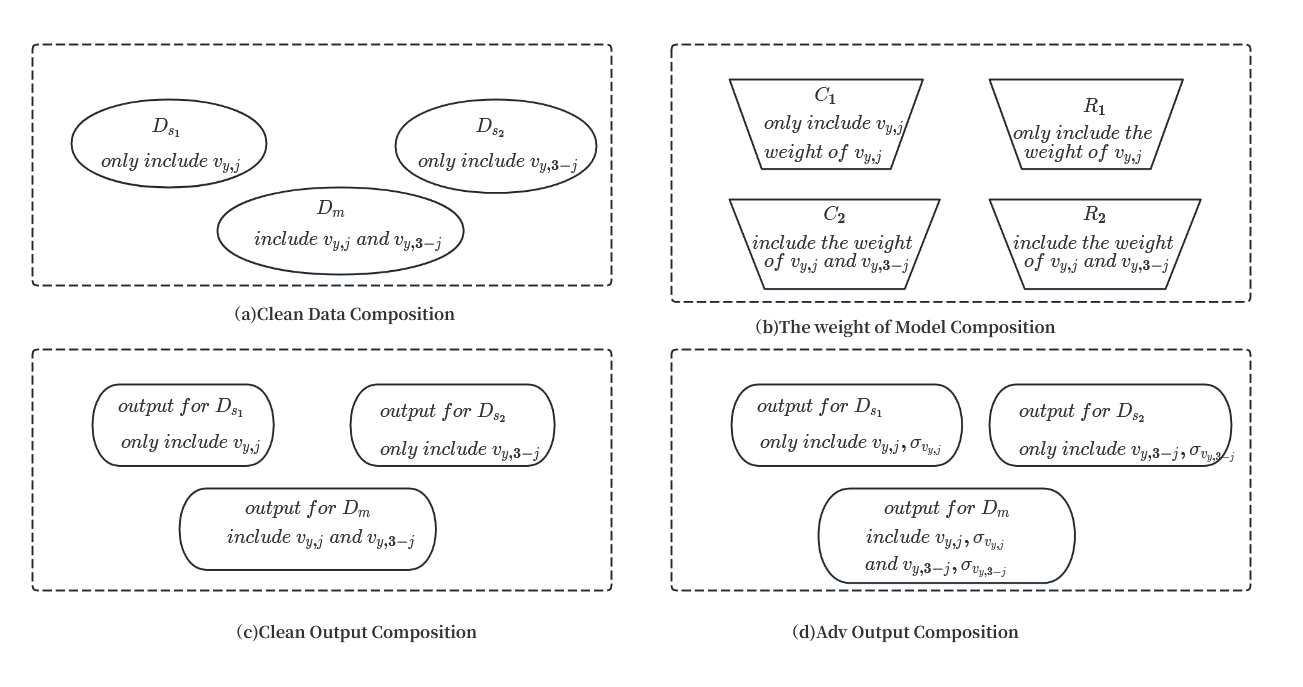}  
	
	\caption{Specific composition of the elements involved in the error calculation}  
	
	\label{fig:my_label}  
	
\end{figure*} 
Examining Equations (11) and (12), we find that learning the \( v_{y,3-j} \) feature is always beneficial in terms of reducing clean training error. However, when the value of \( k_2 \) is large, ${R_{robust}}^{2} - {R_{robust}}^{1}$ will be greater than 0, indicating that learning the \( v_{y,3-j} \) feature can increase the global robust error. Furthermore, it is observed that multi-view data containing the \( v_{y,j} \) feature can hinder the learning of the other \( v_{y,3-j} \) feature in adversarial training.

Specifically, when the condition $(1 - \mu) \left( \frac{e^{\Theta(1)}}{e^{\Theta(1)} + e^{k_{1} \sum_{j' \neq j} \left[ \Theta\left(\frac{k}{d}\right) \right]}} - \frac{e^{2\Theta(1)}}{e^{2\Theta(1)} + e^{(k_{1} + k_{2}) \sum_{j' \neq j} \left[ \Theta\left(\frac{k}{d}\right) \right]}} \right) - 0.5\mu \left( \frac{e^{\Theta(1)}}{e^{\Theta(1)} + e^{k_{2} \sum_{j' \neq j} \left[ \Theta\left(\frac{k}{d}\right) \right]}} \right) \geq 0$ is satisfied, the adversarial training lacks incentive to learn the new feature \( v_{y,3-j} \).

According to \cite{b18}, learned features require a certain amount of adversarial training to purify the dense mixture weights associated with those features. Therefore, in the early stages of learning the \( v_{y,3-j} \) feature, the dense mixture weight $m_{v_{y,3 - {j}}}$ has not been well purified, resulting in larger errors under adversarial perturbations, and thus a larger value of \( k_2 \). Considering the data distribution in the real world, where multi-view proportion \( 1-\mu \) is also relatively large, the probability of this inequality holding is high.As a specific example, taking $\mu = 0.4$, $k_{2}=0.8$, $k_{1}=0.3$, $\Theta(1) = 1$, and ${\sum_{j^{'} \neq j}\left\lbrack \Theta\left( \frac{k}{d} \right) \right\rbrack} = 3$, we have 
$(1 - \mu) \left( \frac{e^{\Theta(1)}}{e^{\Theta(1)} + e^{k_{1} \sum_{j' \neq j} \left[ \Theta\left(\frac{k}{d}\right) \right]}} - \frac{e^{2\Theta(1)}}{e^{2\Theta(1)} + e^{(k_{1} + k_{2}) \sum_{j' \neq j} \left[ \Theta\left(\frac{k}{d}\right) \right]}} \right) - 0.5\mu \left( \frac{e^{\Theta(1)}}{e^{\Theta(1)} + e^{k_{2} \sum_{j' \neq j} \left[ \Theta\left(\frac{k}{d}\right) \right]}} \right) = 0.6 \left( \frac{e^{1}}{e^{1} + e^{0.9}} - \frac{e^{2}}{e^{2} + e^{2.9}} \right) - 0.2 \left( \frac{e^{1}}{e^{1} + e^{2.4}} \right) \geq 0$. This demonstrates that\textbf{ compared to clean training, adversarial training has a high probability of failing to learn some features in single-view data}, which contributes to the decrease in natural generalization of adversarial training.

\section{Proposed Label Design Method}
\subsection{Providing Correct Guidance}
Firstly, we introduce the relevant weights learned by the clean training model for feature \( v \) contained in the labeled data. According to Equation (4), the weights corresponding to feature v learned by clean training consist of two parts. The specific composition of the feature weights \( \omega_{clean-training}^v \) for the clean training model is shown in Equation (13):
\begin{equation}\label{eq:single}{\omega_{clean - training}^{v} = \omega^{v_{pure}} + \omega^{v_{mixture}}}\end{equation}

Where ${\omega_{clean - training}^{v}}$ represents the relevant weights of feature v learned by the clean training model, ${\omega^{v_{pure}}}$ represents the robust and pure part of the relevant weights of feature \( v \), and ${\omega^{v_{mixture}}}$ represents the dense mixture accumulated during gradient descent in clean training.

According to Equation (6), the specific composition of the feature weights ${\omega_{adv - training}^{v}}$ for the adversarial training model is shown in Equation (14):
\begin{equation}\label{eq:single}{\omega_{adv - training}^{v} = \omega^{v_{pure}} + {k\omega}^{v_{mixture}}}\end{equation}
where \( 0< k < 1 \), and the specific value of \( k \) depends on the degree of purification of feature \( v \) by adversarial training.

Ideally, we hope to obtain a robust model whose weights do not contain dense mixtures. The relevant weights of feature \( v \) learned by this ideal model should be as shown in Equation (15):
\begin{equation}\label{eq:single}{\omega_{great - training}^{v} = \omega^{v_{pure}}}\end{equation}

Consider a \( k \)-class classification problem with \( 2k \) features ${v_1, v_2, ..., v2_k}$. Features ${v_{2l-1}}$ and ${v_{2l}}$ correspond to the \( l \)-th class label. In the data distribution:

When belonging to the \( l \)-th class label, we have:

\small \begin{equation} 
	\label{eq:single}{\begin{cases}  
			
			\text{Both } v_{2l-1} \text{ and } v_{2l} \text{ appear with weight } 1,  & \text{accounting for \( 1-u \)} \\  
			
			\text{Only } v_{2l-1}  \text{ appears with weight } 1,  & \text{accounting for \( u/2 \)} \\  
			
			\text{Only } v_{2l} \text{ appears with weight } 1,  & \text{accounting for \( u/2 \)}  
			
\end{cases}}\end{equation}
where \( u/2 \) is the probability of single-view data occurrence, and \( 1-u \) is the probability of multi-view data occurrence.

\textbf{Definition 4.1} (Difference between feature \( v \) in clean and adversarial samples): For  ${\forall~v~ \in ~\mathcal{V}}$, the composition of feature ${v_{advexample}}$ in adversarial samples and feature ${v_{cleanexample}}$ in clean sample data satisfies ${v_{advexample} = v_{cleanexample} + \sigma_{v}}$.

According to 3.2, the clean training model is highly likely to learn two features related to the label, while the adversarial training only learns one feature related to the label. So it can be obtained from equations (6) (13) (14) (15) (16) and Definition 4.1 that $O_{clean - training}^{cleanexample} \approx  O_{great - training}^{cleanexample} =  O_{great - training}^{advexample}$.Therefore, the clean prediction can serve as the label for adversarial samples.On the one hand, this kind of label design can help adversarial training eliminate the dense mixtures in the model features, reduce the impact of perturbations on the model's judgment, and make the model more robust. On the other hand, this label design can assist adversarial training in learning more features, improve the model's judgment of single - view samples during adversarial training, and thus increase the model's natural accuracy.

\subsection{Alleviate the Difficulty of Learning}
As summarized in Section 4.1, the predictions of a clean model can serve as good labels to provide correct guidance for adversarial training. However, the experimental results in RSLAD\cite{b71}, ARD\cite{b69}, and IAD\cite{b70} do not support the viewpoints in Section 3.2. According to their findings, replacing the teacher model in their formulas with a clean model instead of a large robust model leads to unsatisfactory results.

Drawing on the viewpoints in \cite{b73} and \cite{b74}, we found that this may be related to the difficulty of model learning. Perhaps for a model with the capacity of ResNet18, the predictions of the clean training model are too difficult for the adversarial training model to learn. To verify this hypothesis, experiments with different model capacities were designed. Except for the PGDAT experiment, the clean model was set as the teacher model in all experiments. The results of the verification experiments are shown in Table 1:
\begin{table}[]  
	\centering  
	\caption{Impact of model capacity on different methods}  
	\label{tab:model_capacity_effects}
	\begin{subtable}[b]{0.45\textwidth} % 设置子表的宽度和其他参数  
		
		\centering  
		
		\caption{wideresnet34-10}
		\setlength{\tabcolsep}{0pt} % 将列间距设置为 0pt  
		\begin{tabular}{ccc}  
			\toprule 
			{Method}   
			& No Attack & PGD \\ 
			\midrule  
			PGDAT\cite{a16b20} & 86.84 & 56.83 \\  
			ARD\cite{b69} & 88.84 & 58.22 \\   
			IAD\cite{b70} & 87.65 & 60.65 \\ 
			RSLAD\cite{b71} & 87.91 & 60.46 \\  
			LTD\cite{b68a26} & 84.42 & 58.80 \\  
			LBGAT\cite{b50a25} & 88.70 & 59.64 \\ 
			\bottomrule  
		\end{tabular}  
	\end{subtable} 
	
	\bigskip % 增加垂直间距  
	\begin{subtable}[b]{0.45\textwidth} % 设置子表的宽度和其他参数  
		
		\centering  
		
		\caption{resnet18} 
		\setlength{\tabcolsep}{0pt} % 将列间距设置为 0pt
		\begin{tabular}{ccc}  
			\toprule 
			{Method}   
			& No Attack & PGD \\ 
			\midrule  
			PGDAT\cite{a16b20} & 83.92 & 51.71 \\  
			ARD\cite{b69} & 80.11 & 48.60 \\  
			IAD\cite{b70} & 79.82 & 49.71 \\  
			RSLAD\cite{b71} & 86.93 & 46.00 \\  
			LTD\cite{b68a26} & 80.85 & 50.80 \\  
			LBGAT\cite{b50a25} & 77.85 & 45.80 \\  
			\bottomrule  
		\end{tabular} 
	\end{subtable} 
\end{table}

\begin{table*}[]  
	\centering  
	\caption{Comparison with existing work}  
	\footnotesize % 缩小字体
	\label{tab:model_capacity_effects}
	
	\setlength{\tabcolsep}{0pt} % 将列间距设置为 0pt
	\begin{tabular}{cccc} % 手动指定列宽为 2cm
		\toprule
		{Method} & corresponding label &Meaning of T & Measures to reduce learning difficulty \\
		\midrule
		PGDAT & \(f(x'):y\) &  & \\
		TRADES & \(f(x):y \;\;\;  f(x'):f(x)\) & \(T_{\text{big - robust}}(.)\) \\
		ARD & \(f^{\tau}(x):y  \;\;\;  f^{\tau}(x'):T^{\tau}(x)\) & \(T_{\text{big - robust}}(.)\)  &\\
		IAD & \(f^{\tau}(x'):T^{\tau}(x) \;\;\; f^{\tau}(x'):f^{\tau}(x)\) & \(T_{\text{big - robust}}(.)\) & \\
		RSLAD & \(f(x):T(x) \;\;\; f(x'):T(x)\) & \(T_{\text{big - robust}}(.)\) \\
		LTD & \(f(x):T^{\tau}(x)\;\;\; f(x'):f(x)\) & \(T_{\text{clean}}(.)\) & Temperature coefficient
		\\
		LBGAT & \(f(x'):T(x): \;\;\; f(x'):f(x) ~ T(x):y \) & \(T_{\text{clean}}(.)\) & Collaborative training 
		\\
		KD-AT-Auto & \(f(x'):\)The interpolation between \( f^{\tau}(x') \;\;\;and\;\;\; y \) & \(T_{\text{clean}}(.)\) &Temperature coefficient and Interpolation  \\
		HAT & \(f(x):y \;\;\;  f(x'):f(x) \;\;\; f(x'') \):The largest class k other than y & \(T_{\text{clean}}(.)\) &No need to defend against larger perturbations\\
		CKTAT(ours) & \(f(x'):T^{\tau}(x) \;\;f(x'):f(x) \) & \(T_{\text{clean}}(.)\) & Temperature coefficient and Initialization with a clean model\\
		\bottomrule
	\end{tabular}
	
\end{table*}

\begin{table}[t!]  
	
	\caption{Comprehensive ablation experiments of CKTAT}  
	
	\begin{subtable}[t]{0.45\textwidth} % 设置子表的宽度和其他参数  
		
		\centering  
		
		\caption{Removing}  
		\begin{tabular}{ccc}  
			\toprule 
			{Removing}   
			& No Attack & PGD \\ 
			\midrule  
			Clean training model initialization & 82.50 & 55.72 \\  
			\(KL(T^{\tau}(x),S(x^{'})) \) & 82.20 & 55.34 \\   
			\(KL(S(x^{'},S(x)) \) & 86.54 & 54.23 \\ 
			\bottomrule  
		\end{tabular}  
	\end{subtable} 
	\bigskip % 增加垂直间距  
	\begin{subtable}[t]{0.45\textwidth} % 设置子表的宽度和其他参数  
		
		\centering  
		
		\caption{Temperature \( \tau \)} 
		\begin{tabular}{ccc}  
			\toprule 
			{\( \tau \)}   
			& No Attack & PGD \\ 
			\midrule  
			1 & 81.64 & 53.40 \\  
			2 & 81.97 & 53.60 \\  
			3 & 82.82 & 54.71 \\  
			4 & 82.93 & 56.13 \\  
			5 & 83.45 & 56.81 \\  
			6 & 83.09 & 56.41 \\  
			\bottomrule  
		\end{tabular} 
	\end{subtable} 
	\bigskip % 增加垂直间距  
	\begin{subtable}[t]{0.45\textwidth} % 设置子表的宽度和其他参数  
		
		\centering  
		
		\caption{Regularization \( \beta \)} 
		\begin{tabular}{ccc}  
			\toprule 
			{\( \beta \)}   
			& No Attack & PGD \\ 
			\midrule  
			0 & 86.54 & 54.23 \\  
			1 & 85.85 & 54.80 \\  
			2 & 85.10 & 55.20 \\  
			3 & 84.54 & 55.62 \\  
			4 & 83.85 & 55.90 \\  
			5 & 83.65 & 56.30 \\  
			6 & 83.45 & 56.80 \\ 
			7 & 82.85 & 56.89 \\
			\bottomrule  
		\end{tabular} 
	\end{subtable}
\end{table}
\begin{algorithm}  
	\caption{CKTAT: Contrastive Knowledge Transfer based Adversarial Training}  
	\begin{algorithmic}[1]  
		\STATE \textbf{Parameters:}  
		\STATE $D$: Data samples  
		
		\FOR{$i = 1$ to $N_{clean}$}  
		\STATE Sample $(x, y) \sim D$ for a batch  
		\STATE Compute loss based on the basic clean training loss  
		\STATE Update parameters of network $T$ via backpropagation  
		\ENDFOR  
		
		\STATE Initialize the model S with model T
		\FOR{$i = 1$ to $N_1$}  
		\STATE Sample $(x, y_i) \sim D$ for a batch  
		\STATE $x' = \text{attack}(x, y_i)$ to obtain adversarial samples  
		\STATE Compute loss based on Equation (17)  
		\STATE Update parameters of network $S$ via backpropagation  
		\ENDFOR  
	\end{algorithmic}  
\end{algorithm}
Table 1 shows the results of using clean training models with different capacities as teachers. It can be seen that directly using the clean training model as the teacher model on ResNet18 leads to a dual decrease in both the clean accuracy and the robust accuracy of the adversarial training model. However, on WideResNet34 - 10, directly using the clean training model as the teacher results in a dual increase in these two metrics.
Therefore, the experimental conclusions of previous methods can be unified. The predictions of the clean training model are a good label design, but it is necessary to reduce the difficulty of model learning so as to better unleash the potential of the labels.

To reduce the difficulty of model learning, we adopted two measures: using the clean training model for model initialization and scaling the predictions of the clean training model with a temperature coefficient. Using the clean training model for initialization enables the model to avoid relearning multiple features of the samples, thus reducing the learning difficulty. Based on the viewpoints in \cite{b75, b76}, scaling the labels with a temperature coefficient can adjust the difficulty level of model learning.

Thus, the final adversarial training loss function can be obtained, as shown in Equation (17):
\begin{equation}
	\begin{split}
		\label{eq:single}
		\text{loss} = \text{KL}(T^{\tau}(x),S(x^{'})) + \beta\text{KL}(S(x),S(x^{'}))
	\end{split}
\end{equation}
Where \( T(.) \) represents the clean model, \( S(.) \) represents  the adversarial model, \( x \) represents clean samples, \( x^{'} \) represents adversarial samples, and \( \tau \) represents the temperature used for the clean model's output.

\subsection {View on Label Design from Single-view Feature Learning}

All existing work can be explained by the aforementioned feature learning  and learning difficulty reduction. ARD, IAD, and RSLAD use the  predictions of a larger adversarial training model as labels. In terms  of feature learning, a larger adversarial training model has stronger  learning ability, and thus has a higher probability of learning more  features of single - view samples and better removing the perturbation  weights. Therefore, its predictions can provide more correct guidance  than the labels. In terms of learning difficulty, the larger adversarial training model learns fewer features compared to the clean model,  resulting in lower learning difficulty.From the perspective of feature learning, the LTD method uses the  predictions of the clean model as the labels of the adversarial training model for clean samples, enabling the model to learn more single - view features. It indirectly purifies the perturbation weights by aligning  the outputs of the adversarial model for clean samples with those for  adversarial samples. In terms of learning difficulty, it uses a  temperature coefficient, which reduces the learning difficulty.In terms of feature learning, the LBGAT method is consistent with the  method in this paper. In terms of learning difficulty, it conducts  collaborative training of the clean model and the adversarial model,  thereby reducing the learning difficulty.From the perspective of feature learning, KD - AT - Auto utilizes another  adversarial model, which is equivalent to implicit ensemble, increasing  the probability of learning more single - view features. In terms of  reducing learning difficulty, it uses the temperature coefficient and  interpolation method.From the perspective of feature learning, HAT reduces the resistance when  the adversarial model learns features of certain single - view samples,  reducing the probability that the robust error increases when the model  learns such features.

\subsection {Algorithm}
The algorithm includes learning from the clean training model, warm-up training for the adversarial model, and adversarial training with guidance from the clean training model. The final algorithm is described in Algorithm 1:

\section{Experimental Results and Analysis}
In this section, we first conducted a series of comprehensive ablation experiments to fully understand the proposed defense method, and then evaluated its robustness on benchmark datasets in a white-box setting. After that, we analyzed the impact of model size on model parameters and conducted visualization experiments on the features learned by the model. 

\subsection{Comprehensive Ablation Experiments on the Proposed Method}
\begin{table*}[]  
	\centering  
	\caption{Classification Accuracy of Defense Methods under Different Attacks (\%)}  
	\label{tab:model_capacity_effects}
	\begin{subtable}[b]{0.9\textwidth} % 设置子表的宽度和其他参数  
		
		\centering  
		
		\caption{CIFAR-10}  
		\begin{tabular}{ccccccc}  
			\toprule 
			{Defense}   
			& No Attack & PGD & PGD-100 & MIM & CW & AA \\ 
			\midrule  
			PGDAT  & 83.92 & 51.71 & 47.59 & 52.05& 51.18 & 43.77 \\ 
			
			TRADES  & 82.40 & 53.84 & 50.64 & 54.03& 51.47 & 48.72 \\ 
			
			MART  & 83.12 & 55.43 & 53.46 & 57.06 & 51.45 & 48.13 \\  
			
			LBGAT  & 78.93 & 49.57 & 48.01 & 51.88 & 48.97 & 42.68 \\ 
			
			LTD (Clean training model initialization)  & 81.20 & 54.34 & 54.20 & 55.62  & 54.02& 50.56 \\ 
			
			HAT & 82.54 & 56.02 & 55.97 & 57.13& 54.38 & 50.10 \\  
			
			Generalist  & 88.10 & 49.83 & 49.23 & 51.94& 48.85 & 45.99 \\  
			
			CKTAT ($\beta = 0$) & 86.54 & 54.23 & 53.76 & 56.02& 54.07 & 50.01 \\  
			
			CKTAT ($\beta = 6$)  & 83.45 & 56.80 & 56.70 & 58.78 & 55.90& 51.87 \\ 
			\bottomrule  
		\end{tabular}  
	\end{subtable}

	\begin{subtable}[b]{0.9\textwidth} % 设置子表的宽度和其他参数  
		
		\centering  
		
		\caption{CIFAR-100} 
		\begin{tabular}{ccccccc}  
			\toprule 
			{Defense}   
			& No Attack & PGD & PGD-100 & MIM & CW & AA \\ 
			\midrule  
			PGDAT  & 57.42 & 25.58 & 24.62 & 25.16& 23.84 & 21.77 \\  
			
			TRADES  & 55.23 & 27.61 & 26.94 & 27.43& 26.42 & 23.91 \\   
			
			MART  & 50.62 & 28.35 & 27.92 & 28.11 & 27.61 & 24.23 \\  
			
			LBGAT  & 51.24 & 23.20 & 22.73 & 22.95& 23.21 & 20.22 \\   
			
			LTD (Clean training model initialization)  & 54.23 & 27.80 & 27.53 & 27.64& 26.97 & 25.01 \\  
			
			HAT & 59.19 & 26.62 & 26.33 & 26.18& 25.44 & 23.42 \\   
			
			Generalist & 62.58 & 26.24 & 25.89 & 25.92 & 24.25 & 22.15 \\  
			
			CKTAT ($\beta = 0$)  & 59.64 & 27.05 & 26.84 &26.94& 26.04  & 24.54\\
			
			CKTAT ($\beta = 6$)  & 57.20 & 28.51 & 28.50 & 27.60 & 27.40& 25.80 \\ 
			\bottomrule  
		\end{tabular} 
	\end{subtable} 
\end{table*}

\textbf{Removing Terms in CKTAT:}  As shown in Table 3(a), removing the Clean training model initialization  and \(KL(T^{\tau}(x),S(x^{'})) \) terms both resulted in a decrease in the natural accuracy and robust accuracy of CKTAT. However, when the \(KL(S(x^{'},S(x)) \)term was removed, the model's natural accuracy increased, but its robust accuracy rapidly decreased.

\textbf{Sensitivity to Temperature Coefficient ${\tau}$: }The results for different \( \tau \in \left\lbrack 1,6 \right.\rbrack\) are shown in Table 3(b). It can be seen that as \( \tau \) increases, both the model's natural accuracy and robust accuracy increase. The model achieves the best performance when \( \tau \) reaches 5, and then the model's performance gradually deteriorates as \( \tau \) increases.

\textbf{Sensitivity to Regularization Parameter ${\beta}$:}The results for different \( \beta \in \left\lbrack 0,7 \right.\rbrack\) are shown in Table 3(d). If only robustness is considered, then \( \beta=6\) is the best choice. However, if better natural accuracy is desired, a smaller \( \beta\) can be chosen.

\subsection{Robustness Evaluation and Analysis}

It can be seen that on both CIFAR-10 and CIFAR-100 datasets, the proposed CKTAT method showed significant improvement in adversarial accuracy and natural accuracy compared to most of the selected baseline models. CKTAT can adjust clean accuracy and robust accuracy by adjusting the parameter $\beta$. When $\beta=0$, CKTAT's robust generalization is only weaker than LTD and HAT, and its natural generalization is only weaker than Generalist. When $\beta=6$, CKTAT's robust generalization is higher than all selected baseline methods, and its natural generalization is only weaker than Generalist and PGDAT. By utilizing a combination of knowledge transferred from the clean model and some model capability matching techniques, CKTAT achieves better results than multiple baseline models.
\begin{table}[]  
	\centering  
	\caption{Visualization of knowledge points learned by different models}  
	\label{tab:model_capacity_effects}
	
	\centering  
	\renewcommand{\arraystretch}{2.5} 
	\begin{tabular}{cccc}  
		
		&PGDAT   
		& TRADES & CKTAT \\ 
		
		airplane& \begin{minipage}[b]{0.13\columnwidth}
			\centering
			\raisebox{-.5\height}{\includegraphics[width=\linewidth]{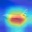}}
		\end{minipage}& \begin{minipage}[b]{0.13\columnwidth}
			\centering
			\raisebox{-.5\height}{\includegraphics[width=\linewidth]{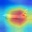}}
		\end{minipage} & \begin{minipage}[b]{0.13\columnwidth}
			\centering
			\raisebox{-.5\height}{\includegraphics[width=\linewidth]{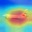}}
		\end{minipage} \\  
		
		cat&\begin{minipage}[b]{0.13\columnwidth}
			\centering
			\raisebox{-.5\height}{\includegraphics[width=\linewidth]{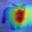}}
		\end{minipage} & \begin{minipage}[b]{0.13\columnwidth}
			\centering
			\raisebox{-.5\height}{\includegraphics[width=\linewidth]{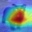}}
		\end{minipage} & \begin{minipage}[b]{0.13\columnwidth}
			\centering
			\raisebox{-.5\height}{\includegraphics[width=\linewidth]{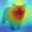}}
		\end{minipage} \\   
		
		bird& \begin{minipage}[b]{0.13\columnwidth}
			\centering
			\raisebox{-.5\height}{\includegraphics[width=\linewidth]{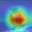}}
		\end{minipage}& \begin{minipage}[b]{0.13\columnwidth}
			\centering
			\raisebox{-.5\height}{\includegraphics[width=\linewidth]{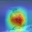}}
		\end{minipage} & \begin{minipage}[b]{0.13\columnwidth}
			\centering
			\raisebox{-.5\height}{\includegraphics[width=\linewidth]{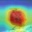}}
		\end{minipage} \\

	\end{tabular}

\end{table}

\subsection{Feature Visualization Experiment}

The visualization experiment was conducted on CIFAR10. As shown in Table 5, CKTAT learns more discriminatory features, followed by TRADES, while PGDAT learns far fewer feature knowledge points. Meanwhile, it can be observed that TRADES learns more features due to the transformation from clean training, while CKTAT learns more features by reducing the difficulty of learning.

\section{Conclusion}
This paper focuses on the label design of adversarial training (AT) and reveals that the decline in generalization ability faced by AT is partly due to the difficulties in learning features of certain single - view samples during adversarial training. It also explores how to establish correct guidance and reduce the learning difficulty through label design. We conducted a systematic analysis of different treatment methods and emphasized that clean training can provide correct guidance for adversarial training, but various means are needed to reduce the learning difficulty during the guidance process.Specifically, we used the predictions of the clean model as correct guidance. Meanwhile, we initialized the model with the clean training model and set a temperature coefficient for the clean model predictions used as labels to reduce the learning difficulty. The experimental results show that adopting this approach in adversarial training enables the model to learn more features and effectively alleviates the decline in the natural generalization ability of adversarial training.

\nocite{langley00}

\bibliography{example_paper}
\bibliographystyle{icml2025}

%%%%%%%%%%%%%%%%%%%%%%%%%%%%%%%%%%%%%%%%%%%%%%%%%%%%%%%%%%%%%%%%%%%%%%%%%%%%%%%
%%%%%%%%%%%%%%%%%%%%%%%%%%%%%%%%%%%%%%%%%%%%%%%%%%%%%%%%%%%%%%%%%%%%%%%%%%%%%%%
% APPENDIX
%%%%%%%%%%%%%%%%%%%%%%%%%%%%%%%%%%%%%%%%%%%%%%%%%%%%%%%%%%%%%%%%%%%%%%%%%%%%%%%
%%%%%%%%%%%%%%%%%%%%%%%%%%%%%%%%%%%%%%%%%%%%%%%%%%%%%%%%%%%%%%%%%%%%%%%%%%%%%%%
\newpage
\appendix
\onecolumn
\section{Appendix A:Data Distribution Definitions and Other Notes}

\subsection{Data Distribution Definitions}
This section references some definitions and notations from \cite{b18}. To simplify the exposition, example choices of certain parameters are used in the definitions. These parameters are primarily functions of \( k \) (the number of classes in the classification problem), such as $P = k^{2},\Gamma = k^{- 1.5},\rho = k^{- 0.01}$, where \( P \) represents the number of image patches, and $\Gamma$ and $\rho$ are parameters used in Definition A.1.

Consider learning a \( k \)-class classification problem on \( P \)-patch inputs, where each patch has dimension \( d \). Here, a "patch" refers to a sub-region of an image, which can be rectangular or square, and its size is determined by the image size and \( P \). Each labeled data point is represented by \( (X, y) \), where the data vector $X = \left( x_{1},x_{2},\ldots,x_{P} \right) \in \left( R^{d} \right)^{P}$ and the data label $y \in \lbrack k\rbrack$.

Assuming \( k \) is sufficiently large, we use the $\mathrm{\Omega},O$ notations to hide polylogarithmic functions of \( k \).

To simplify the problem, we define two features for each class \( j \), represented by unit feature vectors $v_{j,1},v_{j,2} \in R^{d}$. Here, a unit feature vector refers to a vector of length 1 that represents a certain feature in the data. Specifically, as indicated in the fourth point of Definition A.1, the data consists of feature vectors and noise. For simplicity, we assume all features are orthogonal, meaning:

\begin{equation}\begin{split}
		\label{eq:single}{\forall j,{j}^{'} \in \lbrack k\rbrack,\forall l,l^{'} \in \lbrack 2\rbrack,{\parallel v_{j,l} \parallel}_{2} = 1} \\ {and~v_{j,l}\bot v_{{j~}^{'},l^{'}}~~when~(j,~l) \neq ~\left( {j}^{'},l^{'} \right)}
\end{split}\end{equation}

Based on the orthogonality assumption, the features can be extended to the "incoherent" case, where there is no linear relationship between them. The feature set is then defined as:

\begin{equation}\label{eq:single}{\mathcal{V}: = {~\left\{ v_{j,1},~v_{j,2} \right\}}_{j \in \lbrack k\rbrack}~~~~ } \quad  \textbf{\text{the set of all features}}\end{equation}

Next, we cite definitions from \cite{b18} regarding the data distribution of single-view data \( D_s \) and multi-view data \( D_m \), as well as the overall data distribution \( D \) and training dataset \( Z \). Here, \( C_p \) is a global constant, and $s \in \left\lbrack 1,k^{0.2} \right\rbrack$ is a sparsity parameter.

\textbf{Definition A.1} \cite{b18} (Data distribution definition for single-view data \( D_s \) and multi-view data \( D_m \)): Given \( D \in \left\{ D_{m},~D_{s} \right\} \), \( (X,~y)~ \sim ~D \) is defined as follows. First, a label \( y \in \lbrack k\rbrack \) is uniformly randomly selected. Then, the data vector \( X \) is generated as follows:
\begin{enumerate}[label=\arabic*.]  
	
	\item The set of feature vectors used in \( X \) is $\mathcal{V}(X) = \{ v_{y,1}, v_{y,2} \} \cup \mathcal{V}^{\prime}$, where \( \mathcal{V}^{'} \) is a uniformly sampled set of features from $\{ v_{j',1}, v_{j',2} \}_{j' \in [k] \setminus \{ y \}}$, with each feature appearing with probability $\frac{s}{k}$.  
	
	\item For each $v \in \mathcal{V}(X)$, \( C_p \) non-overlapping patches are selected from \( [P] \), denoted as $\mathcal{P}_{v}(X) \subset \lbrack P\rbrack$ (the distribution of these patches can be arbitrary). Define $\mathcal{P}(X) = \cup_{v \in \mathcal{V}(X)}\mathcal{P}_{v}(X)$.  
	
	\item If $D~ = ~D_{s}$ is a single-view distribution, a value $\hat{l} = \hat{l}(X) \in [2]$ is uniformly randomly selected. 
	
	\item For each $v \in \mathcal{V}(X)$ and ${p \in \mathcal{P}}_{v}(X)$, set  $x_{p} = z_{p}v + "noise" \in R^{d}$, where the random coefficient $z_{p} \geq 0	$ satisfies:
	
	In the multi-view distribution case, \( D = D_m \):
	
	${\sum_{{p \in \mathcal{P}}_{v}(X)}{z_{p}}} \in [1, O(1)] \text{ when } v \in \{v_{y,1}, v_{y,2}\},$
	
	${\sum_{{p \in \mathcal{P}}_{v}(X)}{z_{p}~~}} \in ~\left\lbrack \mathrm{\Omega}(\Gamma),~\Gamma \right\rbrack~when~v~ \in ~\mathcal{V}(X)\backslash\left\{ v_{y,1},v_{y,2} \right\},$ 
	
	In the single-view distribution case, \( D = D_s \):
	
	${\sum_{{p \in \mathcal{P}}_{v}(X)}{z_{p}~~}} \in ~\left\lbrack {1,~O(1)} \right\rbrack when~v = v_{y,\hat{{l}}},$
	
	${\sum_{{p \in \mathcal{P}}_{v}(X)}{z_{p}~~}} \in ~\left\lbrack {\rho,~O(\rho)} \right\rbrack when~v = v_{y,3 - \hat{{l}}},$
	
	${\sum_{{p \in \mathcal{P}}_{v}(X)}{z_{p}}} \in \lbrack \mathrm{\Omega}(\Gamma),\Gamma\rbrack when~v \in \mathcal{V}(X)\backslash\{ v_{y,1},v_{y,2} \},$ 
	
	\item For each $p \in \lbrack P\rbrack\backslash\mathcal{P}(X)$, \( x_p \) contains only "noise".
	
\end{enumerate}

\textbf{Definition A.2} \cite{b18} (Definition of the overall data distribution \( D \) and training dataset \( Z \)): The distribution \( D \) consists of data from \( D_m \) ($1-\mu$ portion) and data from \( D_s \) ($\mu$ portion). \( N \) training samples are obtained from \( D \), and the training dataset is represented as $Z~ = ~Z_{m}~ \cup Z_{s}$, where $Z_{m}$ and $Z_{s}$ represent multi-view and single-view training data, respectively. $(X,~y) \sim Z~$ denotes a uniformly randomly sampled \( (X, y) \) from the empirical dataset.

\subsection{Additional Definitions and Notation Annotations}
Let \( M \) denote a neural network model. Let $O_{v_{y,{j}}}^{y}\left( X_{i} \right)$ represent the output of the model \( M \) learned $v_{y,{j}}$ feature-related weights for the clean sample \( X_i \) in the label category \( y \). Let $\sigma_{v_{y,{j}}}$ denote a perturbation in the direction of the dense mixture weight $m_{v_{y,{j}}}$ associated with the feature $v_{y,{j}}$. Let $\hat{X_{i}}$ represent an adversarial sample data (belonging to label category \( y \)) that includes $\sigma_{v_{y,{j}}}$, i.e., $\hat{X_{i}} = X_{i} + \sigma_{v_{y,{j}}}$. Let $F_{i}(X)$ be the model \( M \)'s output for data \( X \) in label category \( i \).

Definitions A.3 and A.4 provide the clean training error and robust training error of model \( M \) on training data:

\textbf{Definition A.3 }(Clean training error of model \( M \) on training data): Given \( N \) training samples, the clean training error of model \( M \) on these \( N \) samples is defined as:

\begin{equation}\label{eq:single}E_{train}^{clean}\left( \mathcal{M} \right)\mathcal{~} = \mathcal{~}\frac{1}{N}{\sum_{i = 1}^{N}\left( 1 - {logit}_{y_{i}}\left( {F,X_{i}} \right) \right)}\end{equation}

where \( y_{i}  \) is the label category of the ith data sample, \( X_{i} \) represents the \( i \)th clean data sample, and ${logit}_{i}\left( {F,X} \right) = \frac{e^{F_{y_{i}}(X)}}{\sum_{j \in ~{\lbrack k\rbrack}}e^{F_{j}(X)}}$.

\textbf{Definition A.4} (Robust training error of model M on training data): Given \( N \) training samples, the robust training error of model \( M \) on these \( N \) samples is defined as:

\begin{equation}\label{eq:single}E_{train}^{adv}\left( \mathcal{M} \right)\mathcal{~} = \mathcal{~}\frac{1}{N}{\sum_{i = 1}^{N}\left( 1 - {logit}_{y_{i}}\left( {F,\hat{X_{i}}} \right) \right)}\end{equation}

where \( y_{i}  \)  is the label category of the ith data sample, \( X_{i} \) represents the ith adversarial data sample.
%%%%%%%%%%%%%%%%%%%%%%%%%%%%%%%%%%%%%%%%%%%%%%%%%%%%%%%%%%%%%%%%%%%%%%%%%%%%%%%
%%%%%%%%%%%%%%%%%%%%%%%%%%%%%%%%%%%%%%%%%%%%%%%%%%%%%%%%%%%%%%%%%%%%%%%%%%%%%%%

\end{document}